\documentclass[journal]{IEEEtran}
\usepackage{latexsym,,amssymb,amsmath,graphicx,epsf,cite,bbm,float}
\usepackage{ifpdf}
\usepackage{epstopdf}

\usepackage{algorithm,algorithmic}
\usepackage{amsmath,amssymb,bm}
\usepackage{amsfonts,dsfont,color,bbm,subcaption}

\def\ninept{\def\baselinestretch{1}}
\ninept

\newcommand{\abs}[1]{|#1|}

\DeclareMathOperator*{\argmin}{arg\,min}

\usepackage{hyperref,amsthm}

\newtheorem{lemma}[]{Lemma}

\newtheorem{corollary}[]{Corollary}

\newtheorem{remark}[]{Remark}

\newtheorem{definition}[]{Definition}

\newtheorem{example}[]{Example}

\begin{document}

\title{Generalized Huber Loss for Robust Learning and its Efficient Minimization for a Robust Statistics} 
\author{\IEEEauthorblockN{Kaan Gokcesu}, \IEEEauthorblockN{Hakan Gokcesu} }
\maketitle

\begin{abstract}
	We propose a generalized formulation of the Huber loss. We show that with a suitable function of choice, specifically the log-exp transform; we can achieve a loss function which combines the desirable properties of both the absolute and the quadratic loss. We provide an algorithm to find the minimizer of such loss functions and show that finding a centralizing metric is not that much harder than the traditional mean and median.
\end{abstract}

\section{Introduction}

Many problems in learning, optimization and statistics literature \cite{poor_book,cesa_book,huberbook,portnoy2000robust} require robustness, i.e., that a model trained (or optimized) be less influenced by some outliers than by inliers (i.e., the nominal data) \cite{hastie2019statistical,huber2004robust}. This approach is extremely common in parameter estimation and learning tasks, where a robust loss (e.g., the absolute error) may be more desirable over a nonrobust loss (e.g., the quadratic error) due to its insensitivity to the large errors. Many penalties for robustness with their particular properties have been proposed in literature \cite{black1996unification,zhang1997parameter}, including parametric formulations to achieve robustness \cite{barron2019general}. In traditional learning approaches, like gradient descent and M-estimation \cite{hampel2011robust}, various losses are commonly used experimentally when designing a system. Normally, the loss metric is provided by the problem formulation itself. However, using a suitably designed loss function can be useful, when the performance evaluation of the resulting learned model is hard to express mathematically. It can be useful in parameter estimation \cite{gokcesu2018density,beck1977parameter} and prediction \cite{gokcesu2016prediction,singer} problems. 

In general, when we have the freedom, instead of using some outlier detection \cite{gokcesu2018anomaly,gokcesu2016nested,gokcesu2019outlier} strategies, it has become important to design loss functions that are intrinsically resistant to outliers. In learning problems, two very commonly used functions are the squared (quadratic) loss, $L(x) = x^2$, and the absolute loss, $L(x)=|x|$. The underlying reasons are the squared loss is strongly convex (hence, has a fast learning rate) and the absolute loss is robust. The squared loss has the disadvantage that it can be dominated by outliers, and when the underlying distribution of the nominal data is heavy-tailed, the efficiency of its minimizer (i.e., the mean) can be poor, i.e., it does not have sufficient distributional robustness.\cite{huber2004robust}. Thus, the estimates may be heavily distorted with some extreme outliers compared to when the outliers are not present. However, the absolute loss does not have these problems, and is robust against arbitrary outliers since their contribution to the estimation is effectively determined by their ordinalities in the data, not their values. Nonetheless, since the quadratic loss is strongly convex, it has fast convergence and learning. Therefore, it is of utmost importance to combine the best of both worlds and create algorithms which are both robust against outliers and have fast convergence near negligible loss.

To create a robust loss with fast convergence, we need to combine the properties of the absolute and the quadratic loss. The most straightforward approach is to use a piecewise function to combine the quadratic and absolute losses where they work the best. As an example, we can straightforwardly use the following function
\begin{align}
L_P(x)=\begin{cases} 
x^2 &, \abs{x}\leq 1\\
\abs{x} &, \abs{x}>1
\end{cases}.
\end{align}
While this function is continuous, the strict cutoff at $1$ may prove to be arbitrary or even useless for certain datasets.
We can bypass this problem by using a free variable as
\begin{align}
L_{P,\delta}(x)=\begin{cases} 
\frac{1}{\delta}x^2 &, \abs{x}\leq \delta\\
\abs{x} &, \abs{x}>\delta
\end{cases}.
\end{align}
While this version is somewhat useful, it is not differentiable, and may prove to be difficult to use in learning tasks. 

To solve the differentiability issue, one can modify the combination as the following, which gives the most popular approach to combine the quadratic and absolute loss functions (i.e., the Huber loss) \cite{huberbook}.
\begin{align}
	L_D(x)=\begin{cases} 
	\frac{1}{2\delta}x^2+\frac{1}{2}\delta &, \abs{x}\leq \delta\\
	\abs{x} &, \abs{x}>\delta
	\end{cases}
\end{align}
Although this function is differentiable, it is not twice differentiable and thus not smooth.

For smoothness, many variants have been proposed. A popular one is the Pseudo-Huber loss \cite{charbonnier1997deterministic}.
\begin{align}
L_{Hp}(x)=\delta\left(\sqrt{1+\frac{x^2}{\delta^2}}\right),
\end{align}
which is $\frac{1}{2\delta}x^2+\delta$ near $0$ and $\abs{x}$
at asymptotes.
While the above is the most common form, other smooth approximations of the Huber loss function also exist \cite{lange1990convergence}.

All in all, the convention is to use either the Huber loss or some variant of it. To this end, we propose a formulation for what we call the Generalized-Huber loss, to encapsulate many different (possible) variants, together with an algorithmic solver. The organization of our paper is the following. In \autoref{sec:general}, we provide the generalized formulation of the Huber loss. In \autoref{sec:convex}, we produce a strictly convex, smooth and robust loss from the generalized formulation. In \autoref{sec:algorithm}, we design an algorithm which minimizes such loss functions. In \autoref{sec:conc}, we finish with further discussions and concluding remarks. 

\section{The Generalized Huber Loss}\label{sec:general}
In this section, we introduce the Generalized-Huber loss. We first start with the definition of a general loss function.
\begin{definition}
	Let $L(\cdot)$ be some loss function such that
	\begin{align*}
	L: \Re\rightarrow\Re, 
	\end{align*}
	where the minimum is at $x=0$, i.e.,
	\begin{align*}
	\argmin_x L(x)=&0,\\
	\min_x L(x)=&L(0).
	\end{align*}
\end{definition}

For a general loss function $L(\cdot)$, we have the following property.
\begin{lemma}
	If $L(\cdot)$ and its first derivative $L'(\cdot)$ are continuous at $x=0$ with positive second derivative (i.e., $L''(0)>0$) and finite higher derivatives, $L(\cdot)$ has a quadratic behavior near $x=0$.
	\begin{proof}
		Near $x=0$, we have
		\begin{align}
			L(x)= L(0)+L'(0)x+\frac{1}{2}L''(0)x^2+o(x^2),
		\end{align}
		from Taylor's expansion. Since $L'(0)=0$ because of continuity and minimum at $0$, we have
		\begin{align}
		L(x)\approxeq L(0)+\frac{1}{2}L''(0)x^2,\label{eq:taylor}
		\end{align}
		near $0$. Hence, it suffices for the loss function to have positive second derivative and finite higher derivatives at zero for its convergence to a quadratic function.
	\end{proof}
\end{lemma}

This result shows that if a loss function and its first derivative are continuous at $x=0$, it has a quadratic behavior for small error. Unfortunately, the absolute loss function does not have a continuous derivative at $x=0$. To solve this, we smooth it with a isotonic/monotonic auxiliary function $f(\cdot)$ \cite{gokcesu2021optimally}. 
\begin{definition}\label{thm:f}
	Let $f(\cdot)$ be some monotone increasing function such that
	\begin{align*}
	\lim\limits_{x\rightarrow\infty}f(x)&=\infty,\\
	\lim\limits_{x\rightarrow -\infty}f(x)&<\infty.
	\end{align*} 
\end{definition}
The auxiliary function $f(\cdot)$ diverges when $x\rightarrow\infty$ and converges when $x\rightarrow-\infty$. Using this auxiliary function, we create the smoothed absolute loss as the following.
\begin{definition}\label{thm:Lg}
	Let the smoothed loss be
	\begin{align*}
	L_G(x)=g(f({x})+f(-x)),
	\end{align*}
	where $g(\cdot)$ is the inverse of $f(\cdot)$, i.e., $f^{-1}(x)$.
\end{definition}
We can see that for a smooth and differentiable $f(\cdot)$ auxiliary function, this loss is also smooth and differentiable. Next, we study its behavior for small and large errors.

\begin{lemma}\label{thm:asymptotes}
	$L_G(\cdot)$ converges to the absolute loss at the asymptotes, i.e.,
	\begin{align*}
	\lim_{\abs{x}\rightarrow\infty}L_G(x)=\abs{x}
	\end{align*}
	\begin{proof}
		When $x$ goes to $\infty$, $f(-x)$ does not diverge, hence,
		\begin{align}
		&L_G(x)\rightarrow g(f(x))=x &&\text{ as } {x\rightarrow\infty},\\
		&L_G(x)\rightarrow g(f(-x))=-x &&\text{ as }{x\rightarrow-\infty},
		\end{align}
		which concludes the proof.
	\end{proof}
\end{lemma}

\begin{lemma}\label{thm:nearzero}
	$L_G(x)$ converges to the quadratic loss near $0$, i.e.,
	\begin{align*}
		L_G(x)\rightarrow ax^2+b &&\text{ as }\abs{x}\rightarrow 0,
	\end{align*}
	where $a=g'(2f(0))f''(0)$ and $b=g(2f(0))$.
	\begin{proof}
		We have
		\begin{align}
			L_G(x)=&g(f_+(x)),\\
			L_G'(x)=&g'(f_+(x))f'_+(x),\\
			L_G''(x)=&g''(f_+(x))[f'_+(x)]^2+g'(f_+(x))f''_+(x).
		\end{align}
		where
		\begin{align}
			f_+(x)\triangleq& f(x)+f(-x),\\
			f'_+(x)=& f'(x)-f'(-x),\\
			f''_+(x)=& f''(x)+f''(-x).
		\end{align}
		Thus, at $x=0$, we have
		\begin{align}
		L_G(0)&=g(2f(0)),\\
		L_G'(0)&=0,\\
		L_G''(0)&=2g'(2f(0))f''(0),
		\end{align}
		since $f_+(0)=2f(0)$, $f'_+(0)=0$ and $f''_+(0)=2f''(0)$. Hence,
		\begin{align}
			L_G(x)\rightarrow g'(2f(0))f''(0)x^2+g(2f(0)) &&\text{ as }\abs{x}\rightarrow 0,
		\end{align}
		from \eqref{eq:taylor}, which concludes the proof.
	\end{proof}
\end{lemma}

\begin{corollary}
    From \autoref{thm:nearzero},	we have for \autoref{thm:f} and \autoref{thm:Lg} the following
	\begin{align*}
	L_G''(0)>0 \iff f''(0)>0.
	\end{align*}
	\begin{proof}
		Since $f(\cdot)$ is monotonically increasing, so is $g(\cdot)$ (i.e., $f^{-1}(x)$). Thus,
		\begin{align*}
		f'(x),g'(x)>0, \forall x
		\end{align*}
		i.e., both derivatives are strictly greater than $0$, which concludes the proof.
	\end{proof}
\end{corollary}

\begin{example}
	When we use a quadratic auxiliary function as
	\begin{align*}
		f(x)=U(x)\left(\frac{x^2}{\delta^2}\right)+1,
	\end{align*}
	where $U(x)$ is the step function. Note that this $f(\cdot)$ is not increasing everywhere and does not have a direct inverse. However, we can use the following pseudo-inverse
	\begin{align*}
		g(x)=\delta\sqrt{x-1}, &&x\geq 1.
	\end{align*}
	Thus, the loss function becomes
	\begin{align*}
		L(x)=g(f(x)+f(-x))=\delta\sqrt{\frac{x^2}{\delta^2}+1}
	\end{align*}
	which is the Pseudo-Huber loss.
\end{example}

This generalized formulation does not guarantee convexity over the whole domain. For convexity to exist, specific functions need to be studied. In the next section, we will study one such function.

\section{A Strictly Convex Smooth Robust Loss}\label{sec:convex}
For convexity, we study the exponential transform. Let 
\begin{align}
	f(x)=e^{ax}+b,
\end{align}
for some $a>0$, hence
\begin{align}
	g(x)=f^{-1}(x)=\frac{1}{a}\log(x-b), &&x> b
\end{align}
and the loss function is
\begin{align}
	L_M(x)=\frac{1}{a}\log(e^{ax}+e^{-ax}+b),
\end{align}
for $b+2>0$. The log-exp transform has a beautiful convexity property.
\begin{lemma}\label{thm:logexp}
	Let $l_i(x)$ for $i\in\{1,\ldots,I\}$ be $I$ convex functions. We have the following convex function $L(\cdot)$
	\begin{align*}
		L(x)\triangleq\log\left(\sum_{i=1}^Ie^{l_i(x)}\right).
	\end{align*} 
	\begin{proof}
		Let $x=\lambda x_1+(1-\lambda)x_2$ for some $0\leq\lambda\leq 1$. We have
		\begin{align}
			\left(\sum_{i=1}^Ie^{l_i(x)}\right)&\leq\left(\sum_{i=1}^Ie^{\lambda l_i(x_1)+(1-\lambda)l_i(x_2)}\right),
		\end{align}
		from the convexity of $l_i(\cdot)$. Setting
		\begin{align}
			a_i&\triangleq e^{\lambda l_i(x_1)},\\
			b_i&\triangleq e^{(1-\lambda) l_i(x_2)},
		\end{align}
		we have
		\begin{align}
			\left(\sum_{i=1}^Ie^{l_i(x)}\right)\leq&\left(\sum_{i=1}^Ia_ib_i\right),\\
			\leq&\left(\sum_{i=1}^Ia_i^{\frac{1}{\lambda}}\right)^\lambda\left(\sum_{i=1}^Ib_i^{\frac{1}{1-\lambda}}\right)^{1-\lambda},
		\end{align}
		from Holder's inequality \cite{hardy1952inequalities}. Thus,
		\begin{align}
			L(x)=&L(\lambda x_1+(1-\lambda)x_2)\\
			=&\log\left(\sum_{i=1}^Ie^{l_i(\lambda x_1+(1-\lambda)x+2)}\right)\\
			\leq&\lambda\log\left(\sum_{i=1}^Ia_i^{\frac{1}{\lambda}}\right)+(1-\lambda)\left(\sum_{i=1}^Ib_i^{\frac{1}{1-\lambda}}\right),\\
			\leq&\lambda\log\left(\sum_{i=1}^Ie^{l_i(x_1)}\right)+(1-\lambda)\left(\sum_{i=1}^Ie^{l_i(x_2)}\right),\\
			\leq&\lambda L(x_1)+(1-\lambda)L(x_2),
		\end{align}
		which concludes the proof.
	\end{proof}
\end{lemma}
This result is intuitive from the smooth maximum \cite{zhang2021dive}.

\begin{remark}
	For $L_M(\cdot)$, its first derivative $L'_M(\cdot)$ and its second derivative $L''_M(\cdot)$; we have the following:
	\begin{align*}
	L_M(x)=&\frac{1}{a}\log(e^{ax}+e^{-ax}+b),\\
	L'_M(x)=&\frac{e^{ax}-e^{-ax}}{e^{ax}+e^{-ax}+b},\\
	L''_M(x)=&\frac{4a+ab(e^{ax}+e^{-ax})}{(e^{ax}+e^{-ax}+b)^2}.\label{eq:L''_M}
	\end{align*}
\end{remark}
\begin{remark}
	From \eqref{eq:L''_M}, we see that for convexity on the whole domain, we need $b\geq0$ (since $a>0$), which is in line with \autoref{thm:logexp}.
\end{remark}

\begin{corollary}
	$L_M(\cdot)$ converges to the absolute loss at the asymptotes from \autoref{thm:asymptotes}, i.e.,
	\begin{align*}
	L_M(x)\rightarrow \abs{x} \text{ as }\abs{x}\rightarrow\infty.
	\end{align*}
\end{corollary}
\begin{corollary}
	$L_M(\cdot)$ converges to the quadratic loss near $0$ from \autoref{thm:nearzero}, specifically
	\begin{align*}
	L_M(x)\rightarrow\left(\frac{1}{a}\log(2+b)+\frac{a}{(2+b)}x^2\right) \text{ as } \abs{x}\rightarrow 0.	
	\end{align*}
	\begin{proof}
		At $x=0$, we have
		\begin{align}
		L_M(0)=&\frac{1}{a}\log(2+b),\\
		L'_M(0)=&0,\\
		L''_M(0)=&\frac{2a}{b+2},
		\end{align}
		for $b+2>0$. The result comes from Taylor's expansion near zero.
	\end{proof}
\end{corollary}

\begin{example}
	For any finite $b$, when $a$ goes to infinity, we have the absolute loss, i.e., $L1$ loss.
\end{example}

\begin{example}
	If $b=-1$, $a=1$, we have direct convergence to $x^2$ near $x=0$. However, while this most straightforwardly combines $\abs{x}$ and $x^2$, it is not convex.
\end{example}

\begin{example}
	If $b=0$ and $a=1$, we have the log-cosh loss \cite{logcosh} translated by $\log(2)$.
\end{example}

\begin{remark}
	When $b\geq 2$, the loss function $L_M(\cdot)$ has the following alternative expression:
	\begin{align*}
	L_M(x)=\frac{1}{a}\log\left(ce^{ax}+\frac{1}{c}\right)+\frac{1}{a}\log\left(ce^{-ax}+\frac{1}{c}\right),
	\end{align*}
	where $c>0$ is such that
	\begin{align*}
	c^2+\frac{1}{c^2}=b.
	\end{align*}
	This formulation is advantages in that it separates the loss function between two asymptotes, which can be straightforwardly used to design different losses with varying asymptotes.
\end{remark}

In the next section, we provide an algorithm to find the minimizer of our loss function.

\section{The Minimizer of Strictly Convex Losses}\label{sec:algorithm}
\begin{algorithm}[!t]
	\caption{Finding the Centralizing Sample Pairs}\label{alg:N}
	\begin{algorithmic}
		\STATE Initialize $I_0=1$ and $I_1=N$.
		\STATE STEP:
		\IF {$I_1=I_0+1$}
		\STATE $x^*_L=x_{I_0}$, $x^*_H=x_{I_1}$
		\STATE Return $x^*_L$ and $x^*_H$
		\ELSIF {$I_1\neq I_0+1$}
		\STATE $I=\left[\frac{I_0+I_1}{2}\right]$, where $[\cdot]$ rounds to the nearest integer
		\STATE Calculate $G=L'\left(x_{I}\right)$
		\ENDIF
		
		\IF {$G=0$}
		\STATE Return the minimizer $x^*=x_{I}$
		\ELSIF {$G>0$}
		\STATE Go to STEP with the update $I_1=I$
		\ELSIF {$G<0$} 
		\STATE Go to STEP with the update $I_0=I$
		\ENDIF
	\end{algorithmic}
\end{algorithm}
Let us have the samples $\{x\}_{n=1}^N$, where we want to minimize the cumulative loss for some function $L_0(\cdot)$, i.e.,
\begin{align}
	\min_xL(x)\triangleq\min_x\sum_{n=1}^{N}L_0(x-x_n).
\end{align}
\begin{remark}
	For the absolute loss, a minimizer is the median:
	\begin{align*}
		\argmin_x\sum_{n=1}^N\abs{x-x_n}=\text{median}(\{x_n\}_{n=1}^N),
	\end{align*}
	i.e., when $\{x_n\}_{n=1}^N$ are ordered, we have
	\begin{align*}
		\text{median}(\{x_n\}_{n=1}^N)=\begin{cases}
		x_{\frac{N+1}{2}}, &\text{N is odd}\\
		\frac{1}{2}(x_{\frac{N}{2}}+x_{\frac{N}{2}+1}), &\text{N is even}
		\end{cases},
	\end{align*} 
	which has $O(N\log N)$ if $\{x_n\}_{n=1}^N$ is unordered.
\end{remark}
\begin{remark}
	For the quadratic loss, the minimizer is the mean:
	\begin{align*}
	\argmin_x\sum_{n=1}^N\abs{x-x_n}^2=\frac{1}{N}\sum_{n=1}^Nx_n,
	\end{align*}
	which has $O(N)$ complexity when $\{x_n\}_{n=1}^N$ is unordered.
\end{remark}
Although both the absolute and the quadratic losses have closed form minimizers, it may not be possible for general loss functions. However, it is possible to find a close minimizer efficiently if the loss $L_0(\cdot)$ is strictly convex as in \autoref{sec:convex}. 

\begin{remark}
	When $L_0(x)$ is strictly convex, we have the following properties:
	\begin{itemize}
		\item $L(x)$ is also strictly convex, hence, it has a unique minimizer $x^*$. 
		\item When the gradient is zero, i.e., $L'(x)=0$ for some $x$; it is the minimizer, i.e., $x=x^*$.
		\item When the gradient is positive, i.e., $L'(x)>0$ for some $x$; it is greater than the minimizer, i.e., $x>x^*$.
		\item When the gradient is negative, i.e., $L'(x)<0$ for some $x$; it is less than the minimizer, i.e., $x<x^*$.
	\end{itemize}
\end{remark}

The properties at this remark are at the core of the algorithm, which get ever so closer to the minimizer with each step of the algorithm. A summary of it is given in \autoref{alg:N}.

\begin{remark}
	The algorithm terminates and returns either the minimizer $x^*$; or, if it was not able to find the minimizer $x^*$, it returns two adjacent sample points $x^*_L$ and $x^*_H$, where the minimizer is such that
	\begin{align*}
		x^*\in(x^*_L,x^*_H).
	\end{align*}
\end{remark}
\begin{remark}
	The runtime of the algorithm is $O(N\log N)$ since the gradient is calculated for $O(\log N)$ times and each calculation takes $O(N)$ time. This linearithmic complexity \cite{sedgewick2011algorithms} is efficient since if the samples $\{x_{n=1}^N\}$ were unordered, ordering them has that much complexity.
\end{remark}

For many applications, finding the centralizing adjacent pair $\{x^*_L,x^*_H\}$ maybe sufficient as in the case of absolute loss for even number of samples. In the absolute loss, the median is an arbitrary minimizer, which is the mean of the centralizing pair by definition. However, each point between that pair is also a minimizer.

If the centralizing pair is not enough and we want to find the minimizer up to a chosen closeness $\epsilon$, we can run a similar algorithm, which is given in \autoref{alg:e}.

\begin{algorithm}[!t]
	\caption{Finding an $\epsilon$-optimal Solution}\label{alg:e}
	\begin{algorithmic}
		\STATE Initialize $x_0=x_L$, $x_1=x_H$ and $\epsilon$.
		\STATE STEP:
		\IF {$x_1-x_0<=2\epsilon$}
		\STATE $x^*_\epsilon=\frac{x_0+x_1}{2}$
		\STATE Return the $\epsilon$-optimal point $x^*_\epsilon$ 
		\ELSIF  {If $x_1-x_0>2\epsilon$} 
		\STATE $\hat{x}=\frac{x_0+x_1}{2}$
		\STATE Calculate $G=L'\left(\hat{x}\right)$
		\ENDIF
		\IF {$G=0$}
		\STATE Return the minimizer $x^*=\hat{x}$
		\ELSIF {if $G>0$}
		\STATE Go to STEP with update $x_1=\hat{x}$
		\ELSIF {$G<0$}
		\STATE Go to STEP with update $x_0=\hat{x}$
		\ENDIF
	\end{algorithmic}
\end{algorithm}

\begin{remark}
	If the algorithm was not able to find the minimizer $x^*$; it returns an $\epsilon$ optimal point $x^*_\epsilon$, which is
	\begin{align*}
		\abs{x^*-x^*_\epsilon}\leq\epsilon
	\end{align*}
\end{remark}
\begin{remark}
	The runtime of the algorithm is $O(N\log(\frac{D}{\epsilon}))$, where $D$ is the separation between $x_L$ and $x_H$, i.e.,
	\begin{align*}
		D\triangleq x_H-x_L.
	\end{align*}
	We have this complexity since the gradient is calculated for $O(\log(\frac{x_H-x_L}{\epsilon}))$ times and each calculation takes $O(N)$ time. $(x_L,x_H)$ can either be $(x^*_L,x^*_H)$ from \autoref{alg:N}, or $(\min_i x_i,\max_i x_i)$ which can be found in $O(N)$ time.
\end{remark}

\section{Discussions and Conclusion}\label{sec:conc}
In this work, we have studied to combine the nice properties of the absolute loss (robustness) and the quadratic loss (strong convexity). Our loss definition is straightforward to use in multivariate case by analyzing each dimension individually. Note that in both absolute and quadratic loss settings, the multidimensional optimization problem reduces to the optimization in each dimension separately.

In literature, there are some nice properties to have for centralizing metrics like the equivariance under scaling, translation, rotation or some other transform \cite{sarle1995measurement}. We point out that every loss function (including ours) that depends on the difference between the parameter $x$ and the samples $x_n$ will be equivariant under translation. However, our loss functions are not equivariant under scaling in contrast to the absolute and quadratic losses. In general, equivariancy under scaling may not even be desirable, especially if $x_n$ are outputs of a non-linear transform. Furthermore, our loss is not equivariant under rotations unlike the quadratic loss. The reason is our loss does not depend on the euclidean distance in the multivariate case (which is needed for rotational equivariancy). Note that the median also is not equivariant. Although there are equivariant extensions like the geometric median \cite{drezner2002weber}, such an enforced intra-dimensional relation may not always be meaningful. Moreover, our loss is not equivariant under arbitrary monotonic transforms unlike the median. Note that the mean is also not equivariant under monotonic transforms. Such a strong property comes with its disadvantages, where the samples values $x_n$ become almost inconsequential and only their ordinality matters. However, this disregard for the values also what makes the median a robust estimator with the highest possible breakdown point (i.e., the most resistant statistic). 

Nonetheless, it is straightforward to make the estimation equivariant. For example, if the input data is whitened at the preprocessing stage, the estimations will be equivariant in rotations. When it is variance normalized (or some other distance measure), it will be equivariant under scalings. Similarly, mean normalization will make it equivariant under translations.

Although the absolute loss is robust, its efficiency decreases substantially when the number of outliers are comparable with the number of nominal data. However, the absolute loss is the limit of convexity. While concave asymptotes can be considered to achieve further robustness, it will eliminate the convexity property, and will require global optimization techniques \cite{gokcesu2021regret}. Such a loss metric will again have a quadratic behavior near $x=0$, will be near linear (i.e., absolute loss) in some intermediate region, and will be the concave function of choice at the asymptotes. Such loss designs can also be found in the literature like the log-linear loss \cite{kim2016robust}.

In conclusion, our work proposes a generalized formulation of the Huber loss. We show that with a suitable function of choice, specifically the log-exp transform; we can achieve a suitable loss function which combines the desirable properties of both the absolute and the quadratic loss. We provide an algorithm to find the minimizer of such loss functions and show that finding a centralizing metric is not that much harder than the traditional mean and median.

\bibliographystyle{IEEEtran}
\bibliography{double_bib}
\end{document}